\documentclass[10pt,twocolumn,letterpaper]{article}

\usepackage{iccv}
\usepackage{times}
\usepackage{epsfig}
\usepackage{graphicx}
\usepackage{amsmath}
\usepackage{amssymb}
\usepackage{subcaption}
\usepackage[utf8]{inputenc}
\usepackage{algorithm}
\usepackage[noend]{algpseudocode}
\usepackage{xcolor}
\usepackage{paralist}
\usepackage{multirow}
\usepackage{amssymb}

\usepackage[pagebackref=true,breaklinks=true,letterpaper=true,colorlinks,bookmarks=false]{hyperref}

\iccvfinalcopy 

\def\iccvPaperID{2677} 

\ificcvfinal\fi
\begin{document}

\title{Decision Explanation and Feature Importance for Invertible Networks}

\author{Juntang Zhuang$^1$, \quad Nicha C. Dvornek $^{2}$,\quad  Xiaoxiao Li$^1$, \quad Junlin Yang$^1$,\qquad James S. Duncan$^{1,2,3}$\\
$^{1}$ Biomedical Engineering, Yale University, New Haven, CT USA  \\ $^{2}$ Radiology \& Biomedical Imaging, Yale School of Medicine, New Haven, CT USA \\  $^{3}$ Electrical Engineering, Yale University, New Haven, CT USA \\
{\tt\small \{j.zhuang; nicha.dvornek; xiaoxiao.li; junlin.yang;james.duncan\}@yale.edu}
}

\maketitle

\begin{abstract}
Deep neural networks are vulnerable to adversarial attacks and hard to interpret because of their black-box nature. The recently proposed invertible network is able to accurately reconstruct the inputs to a layer from its outputs, thus has the potential to unravel the black-box model. An invertible network classifier can be viewed as a two-stage model: (1) invertible transformation from input space to the feature space; (2) a linear classifier in the feature space. We can determine the decision boundary of a linear classifier in the feature space; since the transform is invertible, we can invert the decision boundary from the feature space to the input space. Furthermore, we propose to determine the projection of a data point onto the decision boundary, and define explanation as the difference between data and its projection. Finally, we propose to locally approximate a neural network with its first-order Taylor expansion, and define feature importance using a local linear model. We provide the implementation of our method:  \url{https://github.com/juntang-zhuang/explain_invertible}.
\end{abstract}

\section{Introduction}
Deep learning models have achieved state-of-the-art performance in multiple practical problems, including image classification \cite{lecun2015deep,schmidhuber2015deep}, video processing \cite{baccouche2011sequential} and natural language processing \cite{collobert2008unified}. However, the  black-box nature of the design of most deep learning architectures \cite{castelvecchi2016can} has raised issues such as lack of interpretation and being vulnerable to adversarial attacks \cite{szegedy2013intriguing}. Previous works have found difficulties in recovering images from hidden representations in the neural network \cite{mahendran2016visualizing, dosovitskiy2016inverting}, and it is often unclear what information is being discarded \cite{tishby2015deep}.

Various methods have been proposed to interpret neural networks. The mainstream is to calculate gradient of the loss function $w.r.t$. the input image \cite{zeiler2014visualizing, mahendran2015understanding}. Dosovitskiy et al. proposed up-convolution networks to invert CNN feature maps back to images \cite{dosovitskiy2016inverting}. Another direction for model interpretation is to determine the receptive field of a neuron \cite{zhang2018interpretable} or extract image regions that contribute the most to the neural network decision \cite{zintgraf2017visualizing, kindermans2017learning, kumar2017explaining}. Other works focus on model-agnostic interpretations \cite{sundararajan2017axiomatic, bach2015pixel, lundberg2017unified, li2019efficient}. Different from previous works, we consider explainable neural network models.

The recently proposed invertible network \cite{gomez2017reversible,jacobsen2018revnet, zhuang2019invertible} is able to accurately reconstruct the inputs to a layer from its outputs without harming its classification accuracy. For an invertible classifier, information is only discarded at the final pooling layer and fully-connected layer, while preceding layers preserve all information of the input. This property hints at the potential to unravel the black-box and manipulate data both in the input domain and the feature domain.


In this paper, we introduce a novel method to explain the decision of a network.
We show that an invertible classifier can be viewed as a two-stage model: (1) an invertible transform from the input space to the feature space; (2) a linear classifier in the feature space. For a linear classifier, we can determine the decision boundary and explain its prediction; using the invertible transform, we can determine the corresponding boundary and explanation in the input space. 

After determining the projection onto the decision boundary, we perform Taylor expansion around the projection, to locally approximate the neural net as a linear function. Then we define the importance using the same method as in linear classifier cases. 

Our main contributions can be summarized as:
\begin{compactitem}
    \item We explicitly determine the decision boundary of a neural network classifier and explain its decision based on the boundary. 
    \item We use Taylor expansion to locally approximate the neural net as a linear function and define the numerical importance of each feature as in a linear classifier.
\end{compactitem}
\vspace{-0.4cm}
\section{Invertible Networks}
The network is composed of different invertible modules, followed by a global average pooling layer and a fully-connected layer. Details for each invertible module are described in the following sections.

\subsection{Invertible Block}
\begin{figure}[!h]
\begin{minipage}[t]{0.50\linewidth}
    \centering
    \includegraphics[height=1.5cm]{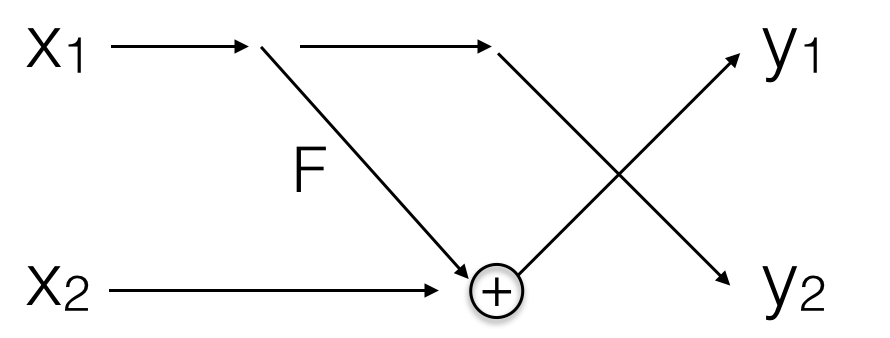}
    \caption{\small{Structure of the invertible residual block.}}
    \label{fig_rev_block}
\end{minipage} \hfill
\begin{minipage}[t]{0.4\linewidth} 
    \centering
\includegraphics[angle=90,width=3cm]{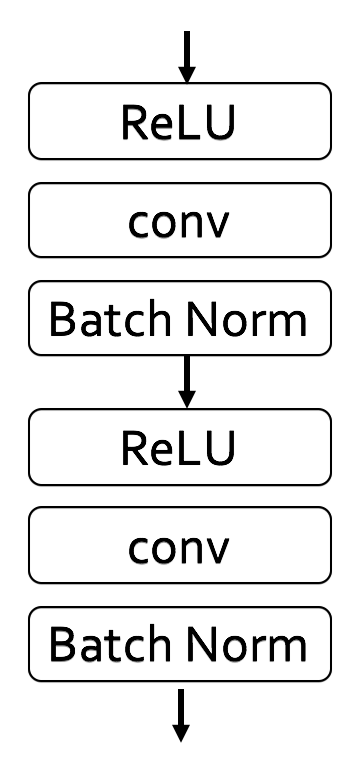}
    \caption{
    \small{An example of $F$ in the invertible block.}
    }
    \label{F_block}
\end{minipage}        
\end{figure}

An invertible block serves a similar role as a building block like a residual block, except it is invertible. For the invertible block in Fig.~\ref{fig_rev_block}, we follow the structure of the reversible block in \cite{gomez2017reversible}. The input $x$ is split into two parts $x_1$ and $x_2$ by channel, such that $x_1$ and $x_2$ have the same shape.  

Corresponding outputs are $y_1$ and $y_2$ with the same shape as the input. $F$ represents some function with parameters to learn, and $F$ can be any continuous function whose output has the \textbf{same} shape as input; an example of $F$ is shown in Fig.~\ref{F_block}. $F$ can be convolutional layers for 2D inputs and FC layers for 1D inputs. The forward pass and inversion is calculated as:
\begin{equation} \label{eq_rev_forward}
   \left\{
                \begin{array}{ll}
                  y_1 = x_2 + F(x_1)\\
                  y_2 = x_1\\
                \end{array}
    \right.
    \left\{
                \begin{array}{ll}
                  x_1 = y_2\\
                  x_2 = y_1 - F(x_1)\\
                \end{array}
    \right.
\end{equation}
\vspace{-0.1cm}
\vspace{-0.5cm}
\subsection{Invertible Pooling with 2D Wavelet Transform}
\begin{figure}[!h]
    \centering
    \includegraphics[width=0.8\linewidth]{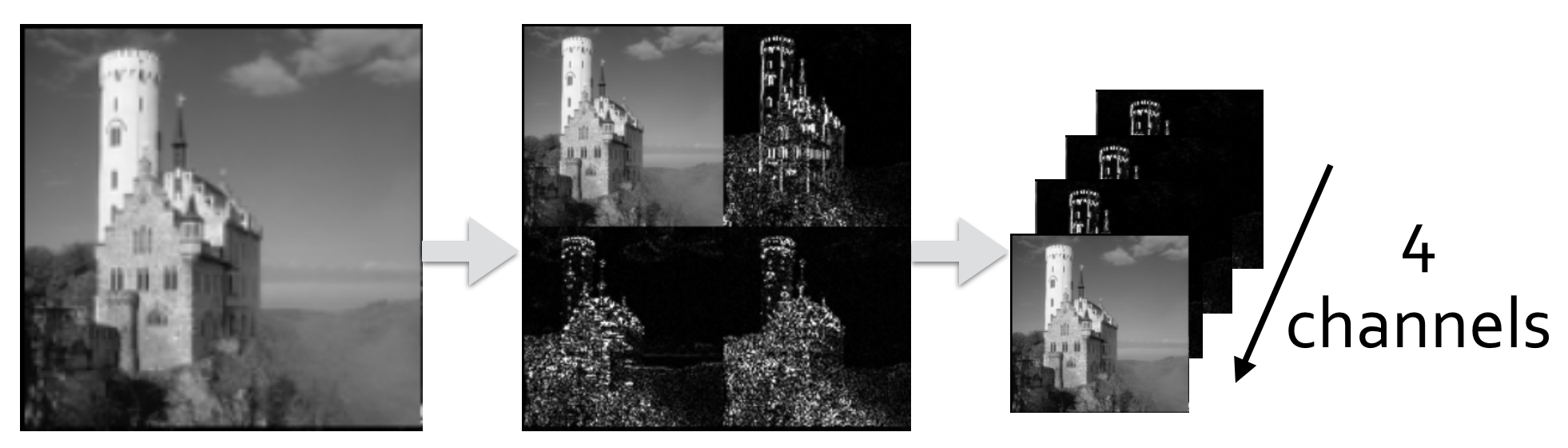}
    \caption{\small{2D wavelet transform as an invertible pooling.}}
    \label{fig:pool}
\end{figure}

An invertible pooling can halve the spatial size of a feature map, and reconstruct the input from its output.
We use the 2D wavelet transform at level 1 as shown in Fig. \ref{fig:pool}. Each channel of a tensor is a 2D image. A 2D image is transformed into 4 sub-images whose height/width is half of the original image. Four sub-images are stacked into 4 channels. The inversion can be calculated by the inverse 2D wavelet transform.
\subsection{Inverse of Batch Normalization}
The forward pass and inverse of a batch normalization layer are listed below:
\begin{equation} \label{eq_bn_forward}
y = \frac{x- \textbf{E}(x)}{\sqrt{\textbf{Var}(x)+\epsilon}} \gamma + \beta,\ \ \ x = \frac{y-\beta}{\gamma} \sqrt{\textbf{Var(x)}+\epsilon} + \textbf{E}(x)
\end{equation}
where $\gamma$ and $\beta$ are parameters to learn, $\textbf{E}(x)$ and $\textbf{Var}(x)$ are approximated as the sample-mean and sample-variance respectively, and all operations are channel-wise. 

\subsection{Linear Layer}
\label{sec_linear}
The feature space usually has a high dimension compared to the number of classes for final prediction. The mapping from high-dimension to low-dimension is typically performed with an average pooling and a fully-connected (FC) layer in a convnet. These two steps combined is still a linear transform and can be denoted as:
\begin{equation}
    y = A B z =  W z, where\ \ W = AB
\end{equation}
where $z$ is a $C\times h \times w$ feature vector reshaped to 1D, $h,w$ are spatial sizes, and $C$ is the channel number; $B$ is a block-wise constant matrix of size $C \times Chw$, representing the average pooling operation; $A$ is the weight of a FC layer with size $K \times C$, where $K$ is the number of classes; and $W=AB$ combines the two steps into 1 transform matrix.
\vspace{-0.3cm}
\begin{figure}[!htb]
    \centering
    \includegraphics[width=0.7\linewidth]{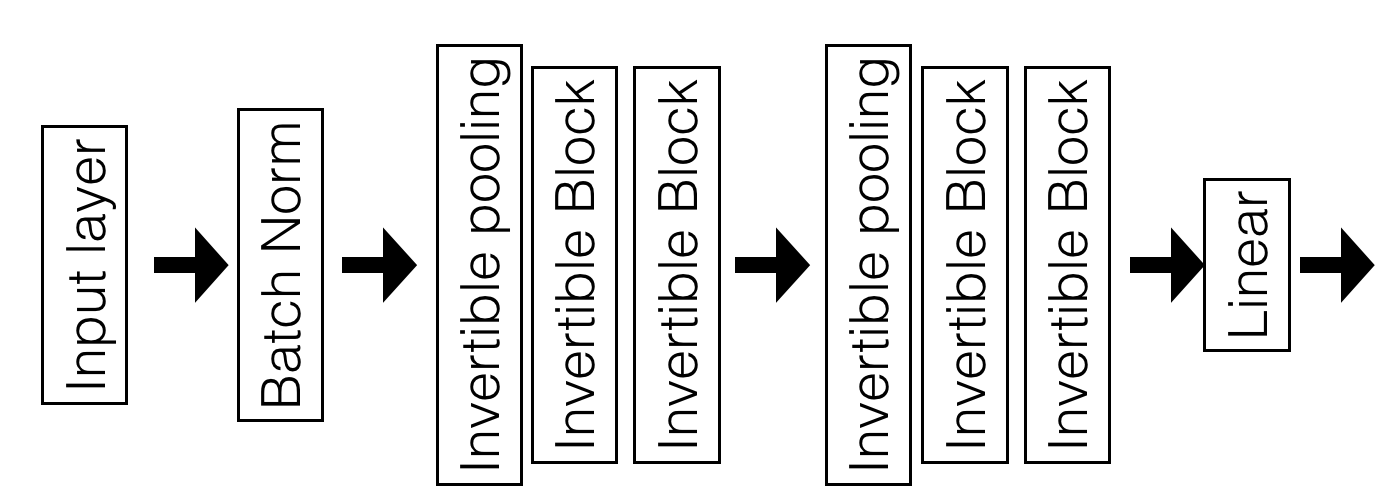}
    \caption{Structure of an invertible network.}
    \label{fig_network}
\end{figure}
\vspace{-0.4cm}
\subsection{Structure of Invertible Network}
The structure of a classification network is shown in Fig. \ref{fig_network}. The network is invertible because its modules are invertible. The input image is fed into a batch normalization layer followed by an invertible pooling layer. The invertible pooling layer increases the channel number by 4 and is essential to make the tensor have an even number of channels in order to keep the same shape for $x_1$ and $x_2$ as in formula \ref{eq_rev_forward}.

The network is divided into stages, where an invertible pooling layer connects two adjacent stages. Within each stage, multiple invertible blocks are stacked. The output from the final stage is fed into a linear layer defined in Sec.~\ref{sec_linear}. The probability of current data belonging to a certain class is calculated as a $softmax$ of the logits.  

\subsection{Reconstruction Accuracy of Inversion}
We build an invertible network of 110 layers. We train the network on the CIFAR10 dataset \cite{krizhevsky2009learning}, and reconstruct the input image from the output of the final invertible block. Results are shown in Fig. \ref{fig:test_recon_accuracy}. The $l_2$ distance between reconstruction and input is on the order of $10^{-6}$, validating the accuracy of inversion.

\section{Interpret Model Decision}
\subsection{Notations of Network}
\label{sec:notation}
The invertible network classifier can be viewed as a two-stage model: 
\begin{equation} \label{eq_whole_net}
    t = T(x), \;  y = Class(t)
\end{equation}
(1) The data is transformed from the input space to the feature space by an invertible function $T$. (2) Features pass through a linear classifier $Class$, whose parameters $W$ and $b$ are defined in Sec.~\ref{sec_linear}.  

\vspace{-0.2cm}
\begin{figure}[h]
\begin{minipage}[b]{0.5\linewidth}
    \centering
    \includegraphics[width=\linewidth]{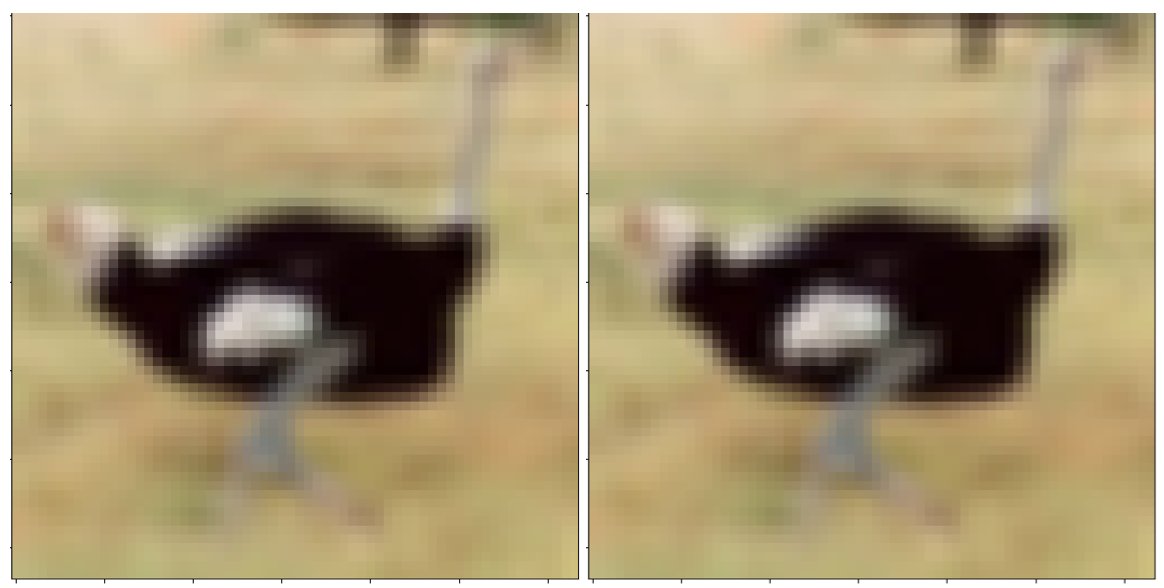}
    \caption{\small{
    From left to right: input image, reconstructed image from outputs of last invertible block.}}
    \label{fig:test_recon_accuracy}
\end{minipage} \hfill
\begin{minipage}[b]{0.45\linewidth}
    \centering
    \includegraphics[ height=1.8cm]{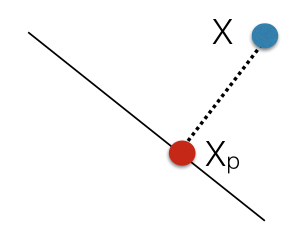}
    \caption{\small{
    For a linear classifier, $X_p$ is the projection of $X$ onto the decision plane, and the vector ($X_p,X$) is the explanation for decision.
    }}
    \label{fig:linear_explanation}
\end{minipage}
\end{figure}
\vspace{-0.2cm}

The operation of $y=Class(t)$ is defined as:
\begin{equation} \label{eq_linear}
   \left\{
                \begin{array}{ll}
                  y_k = \langle \vec{t},\vec{w_k} \rangle + b_k,\  \ k=1,2,...K \\
                  \textbf{P}(k \vert x) = \frac{exp(y_k)}{\sum_{i=1}^K exp(y_i)}\\
                \end{array}
              \right.
\end{equation}
where $\vec{w_k}$ is the weight vector for class $k$, also is the $k$th row of $W$ in Sec.~\ref{sec_linear}; $b_k$ is the bias for class $k$; $\langle \cdot,\cdot \rangle$ is the inner-product operation and $K$ is the total number of classes.

\subsection{Determine the Decision Boundary} 
\label{sec_determine_boundary}
Note that on the decision boundary probabilities of two classes are the same. Using the same notation as in formula \eqref{eq_whole_net} and \eqref{eq_linear}, the decision boundary between class $i$ and $j$ in the feature domain is:
\begin{equation} \label{eq_plane}
\langle \vec{t},\vec{w_i} \rangle + b_i = \langle \vec{t},\vec{w_j} \rangle + b_j
\end{equation}
The solution $\vec{t}$ to formula \eqref{eq_plane} lies on a high-dimensional plane, and can be solved explicitly.

Since $T$ is invertible, we can map the decision boundary from the feature space to the input domain. 

\subsection{Model Decision Interpretation}
\label{sec:interpret_boundary}
\subsubsection{Interpret linear models}
We first consider a linear classifier for a binary problem as in Fig. \ref{fig:linear_explanation}. For a data point $X$ and its projection $X_p$ onto the decision boundary, $X_p$ is the \textit{nearest} point to $X$ on the boundary; the vector ($X_p,X$) could be regarded as the explanation for the decision, as shown below:
\begin{equation}
    Explanation = X - X_p
\end{equation}

\vspace{-0.5cm}
\subsubsection{Interpret non-linear model}
\begin{figure}[!htb]
    \centering
    \includegraphics[width=0.65\linewidth]{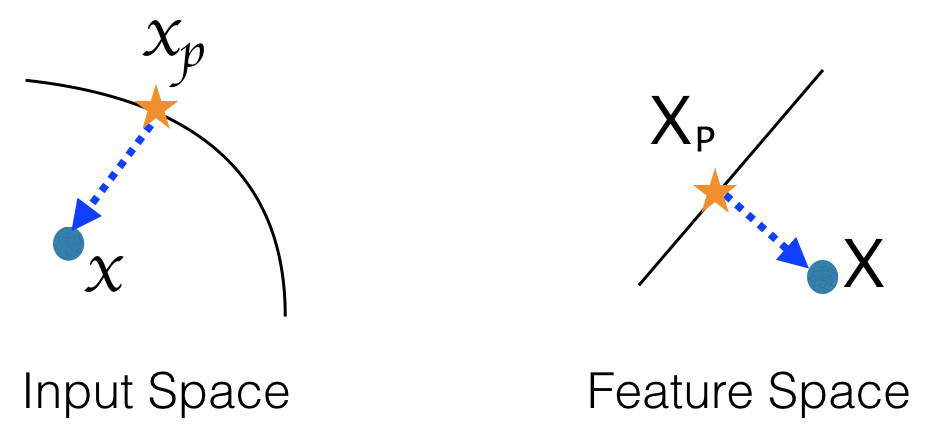}
    \caption{\small{Explanation of invertible non-linear classifiers. Left figure is the input space, right figure is the feature space. Black line is the decision boundary, $X_p$ is the projection of $X$ onto the decision boundary. Vector $( X,X_p )$ is perpendicular to the boundary in feature space. Dashed vector can be viewed as the explanation for model decision.}}
    \label{fig:explain}
\end{figure}

The last layer of a neural network classifier is a linear classifier, and we can calculate $X_p$ from $X$ as in linear case. With invertible networks, we can find their corresponding inputs, denoted as $T^{-1}(X)$ and $T^{-1}(X_p)$ respectively, where $T$ is the transform function as in equation \eqref{eq_whole_net}. Vector ($T^{-1}(X),T^{-1}(X_p)$) is the explanation in the input domain.   
The explanation can be denoted as:
\vspace{-0.1cm}
\begin{equation}
    Explanation = T^{-1}(X) - T^{-1}(X_p)
\end{equation}
\vspace{-0.1cm}
where $X = T(x)$ is the point in the feature space, corresponding to $x$ in the input space; $X_p$ is the projection of $X$ onto the boundary in the feature space; and $x_p$ is the inversion of $X_p$, as shown in Fig.~\ref{fig:explain}.
\vspace{-0.3cm}
\vspace{-0.8cm}
\subsubsection{Feature importance}
\paragraph{Linear case}
For a linear model, ignoring the bias, the function for log-probability is:
\vspace{-0.2cm}
\begin{equation}
    f(x) =\sum_i^d w_i (x_i - x_{p,i})
\end{equation}
where $d$ is the dimension of $x$; $x_p$ is the projection of $x$ onto the boundary, which is also the nearest point on the boundary; and $w_i$ is the weight for dimension $i$. 

The explanation in dimension $i$ is $x_i - x_{p,i}$; the contribution to $f(x)$ is $w_i (x_i - x_{p,i})$. Therefore, we define $\vert w_i (x_i - x_{p,i}) \vert$ as the importance of feature $i$ for data $x$. 
\vspace{-0.2cm}
\paragraph{Non-linear case}
We use Taylor expansion around $x_p$ to approximate the neural network with a linear model locally:
\begin{equation}
    f(x) = f(x_p) + \nabla f(x_p)^T (x - x_p) + O(\vert \vert x-x_p \vert \vert^2_2)
\end{equation}
For a local linear classifier, the importance of each feature is:
\begin{equation}
    Importance = \vert \nabla f(x_p)\odot (x-x_p) \vert
\end{equation}
where $\odot$ is the element-wise product, and $Importance$ is a vector with the same number of elements as $x$.

\section{Experiments}
\subsection{Decision Boundary Visualization}
\label{visualize_boundary}
For a $d$-dimensional input space, the decision boundary is a $(d-1)$-dimensional subspace. For the ease of visualization, we perform experiments on a 2D simulation dataset, whose decision boundary is a 1D curve. 

\begin{figure}[h] 
        \centering
        \begin{subfigure}[b]{0.48\linewidth}
            \centering
            \includegraphics[width=0.7\linewidth]{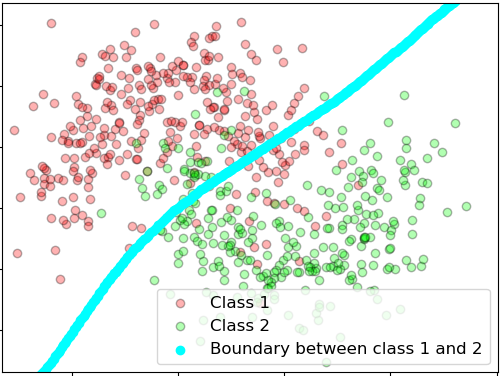}
            \caption{\footnotesize{Input domain with 1\% of the training data.}}%
        \end{subfigure}
        \begin{subfigure}[b]{0.48\linewidth}   
            \centering             \includegraphics[width=0.7\linewidth]{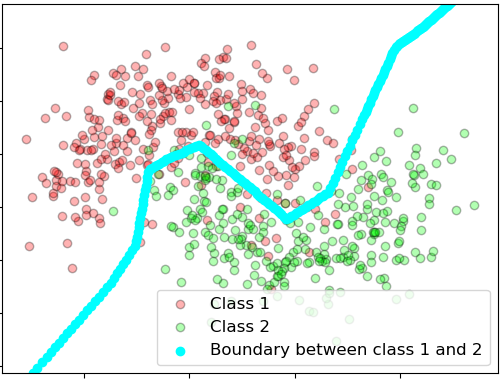}
            \caption{\footnotesize{Input domain with 100\% of training data.}}%
        \end{subfigure}
        \\
        \begin{subfigure}[b]{0.48\linewidth}
            \centering
            \includegraphics[width=0.7\linewidth]{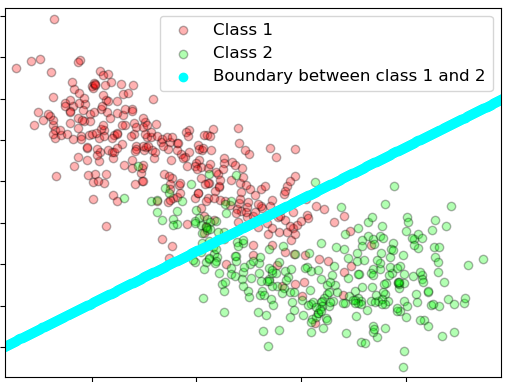}
            \caption{\footnotesize{Feature domain with 1\% of training data.}}%
        \end{subfigure}
        \begin{subfigure}[b]{0.48\linewidth}   
            \centering             \includegraphics[width=0.7\linewidth]{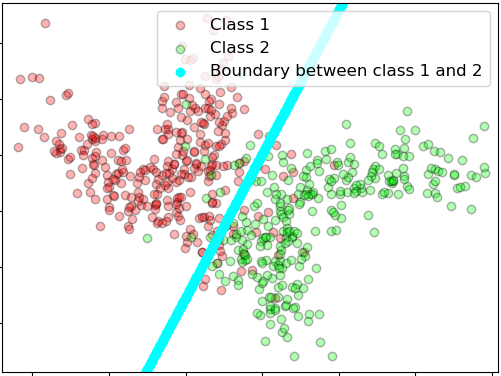}
          \caption{\footnotesize{Feature domain with 100\% training data.}}%
        \end{subfigure}
        \caption{
        \small{Visualization of the decision boundary varying with the size of training set on a 2D toy dataset. Top row shows results in input domain, and bottom row shows results in feature domain. Columns left (right) shows training with 1\% (100\%) of data.}
        }
        \label{fig:datasetsize_boundary}
\end{figure}

The data points for two classes are distributed around two interleaving half circles. As shown in Fig.~\ref{fig:datasetsize_boundary}, two classes are colored with red and green. The decision boundary is colored in blue. We visualize the decision boundary in both the input domain and the feature domain. 

Visualization of the decision boundary can be used to understand the behavior of a neural network.
We give an example to visualize the influence of training set size on the decision boundary in Fig. \ref{fig:datasetsize_boundary}. From left to right, the figure shows the decision boundary when training with 1\% and 100\% of data, respectively. As the number of training examples increases, the margin of separation in the feature domain increases, and the decision boundary in the input domain gradually captures the moon-shaped distribution. Furthermore, the decision boundary can be used to determine how the network generalizes to unseen data.
\vspace{-0.1cm}
\vspace{-0.5cm}
\subsection{Feature Importance}
We validated our proposed feature importance method on a simulation dataset using $make\_classification$ in $scikit-learn$. We created a 2 class, 10-dimensional dataset, of which only 3 dimensions are informative. 

\begin{figure}[!htb]
    \centering
    \includegraphics[width=0.8\linewidth]{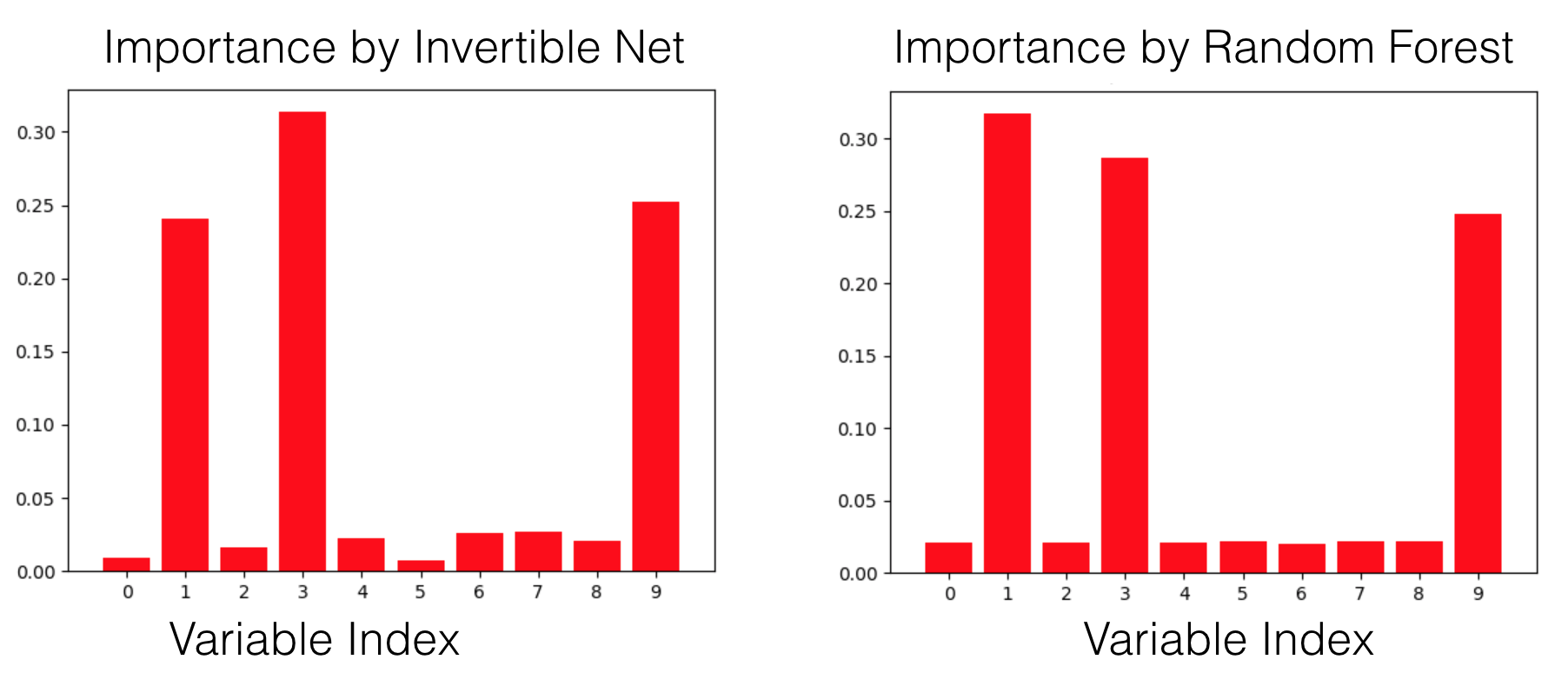}
    \caption{\small{Left: feature importance from invertible network. Right: feature importance from random forest.}}
    \label{fig:importance}
\end{figure}

We train an invertible network and computed the importance of each dimension. Results are shown in Fig.~\ref{fig:importance}. An oracle model should give equal importance to 3 informative variables (indexed by 1, 3 and 9), while set 0 to other variables. Our invertible network successfully picks informative features, and generates feature importance comparable to random forest. Both models select the correct features.
\vspace{-0.0cm}
\vspace{-0.0cm}
\subsection{Explain a Convolutional Invertible Network}
We train a convolutional invertible classifier, achieving over 99\% accuracy on the MNIST test set. 
For an input image, we select classes with the top 2 predicted probabilities, determine the decision boundary between these two classes as in Sec.~\ref{sec_determine_boundary}, calculate the projection onto the boundary, and interpolate between the input image and its projection onto the boundary in the feature domain. 

\begin{figure}[!htb]
    \centering
    \includegraphics[width=0.7\linewidth]{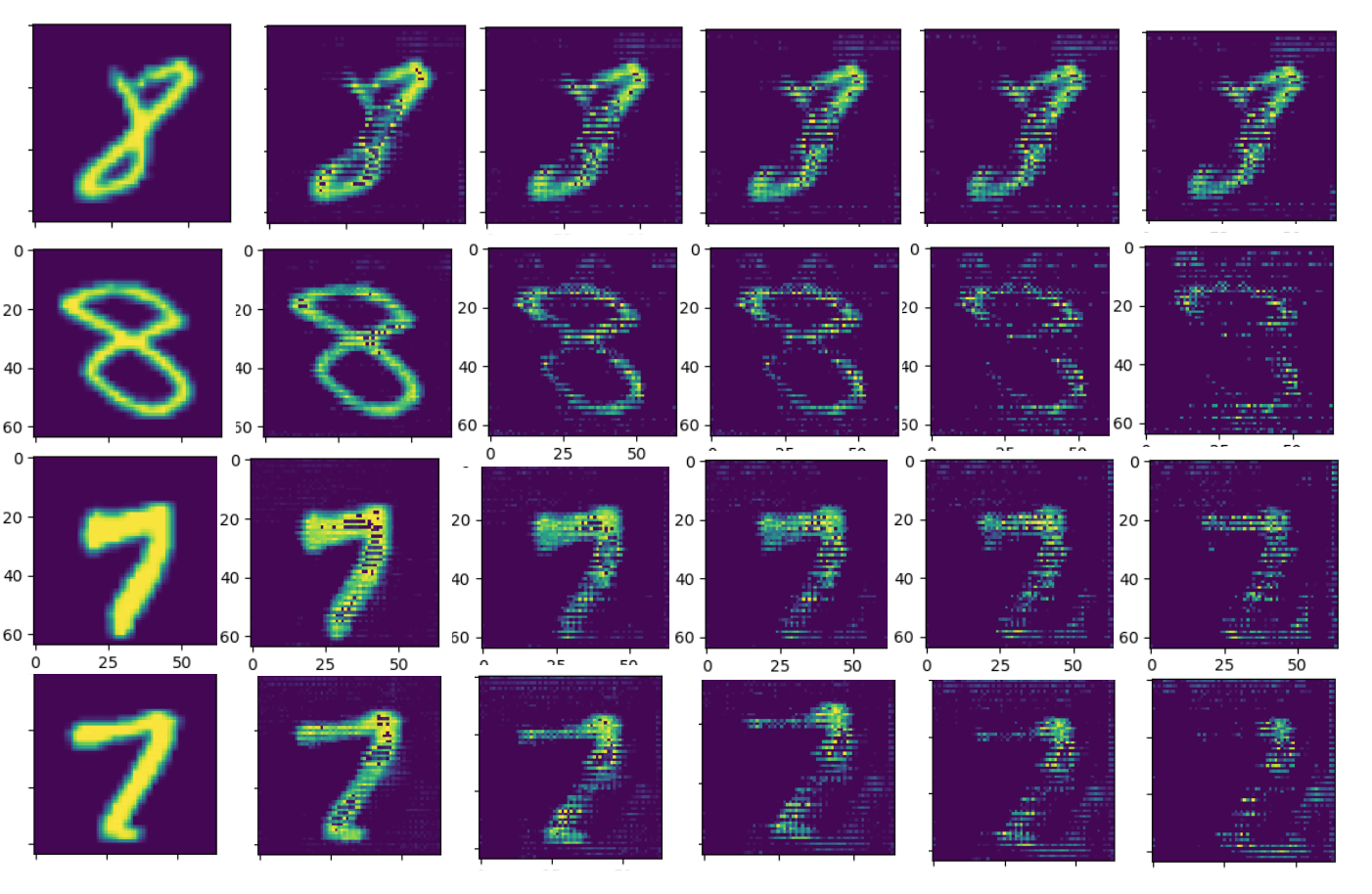}
    \vspace{-0.1cm}
    \caption{\small{Interpolation (performed in the feature domain) between input (left most) and its projection on the boundary (right most). Top two rows shows 8 transforms to 3, bottom two rows show 7 to 2.}}
    \label{fig:mnist}
\end{figure}

Results are shown in Fig.~\ref{fig:mnist}. Note that for each row, only one input image (left most) is provided; the projection (right most) is calculated from the model, instead of searching for a nearest example in the dataset. So the projection demonstrates the behavior of the network. As discussed in Sec.~\ref{sec:interpret_boundary}, the difference between a data point and its projection onto the boundary can be viewed as the explanation. For example, for an image of 8, its left half vanishes in the interpolation, which explains why it's not classified as 3; for an image of 7, a bottom line appears in the interpolation, which explains why it's not classified as 2.
\vspace{-0.2cm}
\vspace{-0.6cm}
\section{Conclusion}
\vspace{-0.1cm}
We propose a method to explicitly determine the decision boundary of an invertible neural network classifier and define the explanation for model decision and feature importance. We validate our results in experiments, and demonstrate that the transparency of invertible networks has great potential for explainable models. 

{\small
\bibliographystyle{ieee}
\bibliography{egbib}
}

\end{document}